%% file: main.tex
\documentclass[conference, letterpaper, 10pt]{IEEEtran}
\usepackage{cite}
\usepackage{amsmath,amssymb,amsfonts}
\usepackage{algorithmic}
\usepackage{graphicx}
\usepackage{textcomp}
\usepackage{xcolor}
\def\BibTeX{{\rm B\kern-.05em{\sc i\kern-.025em b}\kern-.08em
    T\kern-.1667em\lower.7ex\hbox{E}\kern-.125emX}}
    
\usepackage{listings}
\definecolor{myblue}{HTML}{003399}
\definecolor{mygray}{HTML}{555555}
\usepackage{tikz}
\usetikzlibrary{arrows, calc, matrix, positioning}
\usepackage{multirow}
\usepackage{hyperref}
\hypersetup{
    colorlinks=true,
    citecolor=black,
    linkcolor=black,
    urlcolor=myblue
}
\usepackage{tipa}

\begin{document}

\title{
Multi-GPU SNN Simulation\\with Static Load Balancing
}

\def \CSD {CSD\,-\,UOC}
\def \ICS {FORTH\,-\,ICS}

\author{
\IEEEauthorblockN{1\textsuperscript{st} Dennis Bautembach}
\IEEEauthorblockA{\textit{\ICS{} \& \CSD{}}\\denniskb@ics.forth.gr}
\and
\IEEEauthorblockN{2\textsuperscript{nd} Iason Oikonomidis}
\IEEEauthorblockA{\textit{\ICS{}}\\oikonom@ics.forth.gr}
\and
\IEEEauthorblockN{3\textsuperscript{rd} Antonis Argyros}
\IEEEauthorblockA{\textit{\ICS{} \& \CSD{}}\\argyros@ics.forth.gr}}

\maketitle

\begin{abstract}
We present a SNN simulator which scales to millions of neurons, billions of synapses, and 8 GPUs. This is made possible by 1) a novel, cache-aware spike transmission algorithm 2) a model parallel multi-GPU distribution scheme and 3) a static, yet very effective load balancing strategy. The simulator further features an easy to use API and the ability to create custom models. We compare the proposed simulator against two state of the art ones on a series of benchmarks using three well-established models. We find that our simulator is faster, consumes less memory, and scales linearly with the number of GPUs.
\end{abstract}


\input{1_intro}
\input{2_related}
\input{3_method}
\input{4_results}
\input{5_conclusion}

\section*{Acknowledgments}
We thank Sergiu Oprea for his inputs on our algorithm and the authors of BSim and NeuronGPU for their assistance with the benchmarks.

\bibliographystyle{IEEEtran}
\bibliography{manual, mendeley}

\end{document}

%% file: 1_intro.tex
\section{Introduction}
Spiking Neural Networks (SNNs), first formalized in 1997~\cite{maass1997}, have experienced a renaissance in recent years due to the rise in popularity of Deep Learning and the widespread availability of GPGPU hardware. While we are still waiting for the breakthrough that will let SNNs overtake 2nd generation ANNs, the research community remains highly active, working to improve performance, biological fidelity, support for complex models and topologies, and user friendliness. Our own prior work~\cite{bautembach2020} surpassed the state of the art on a couple of these criteria.

We believe that SNN simulation is fundamental to SNN design: faster simulation means faster iteration and thus quicker progress for the field as a whole. Not only speed is important, but so is size, especially considering that we are currently limited to simulating a modest 1\% of a rat's visual cortex~\cite{Knight2018}. While gains are still to be made from algorithmic and data-structural improvements, we must solve multi-GPU (and eventually multi-node) scaling. Towards this goal we present a SNN simulator called ``Spice'' (\textipa{/spaIk/}) which scales to millions of neurons, billions of synapses, and 8 GPUs.\footnote{We use ``GPU'' to refer to an entire GPU PCIe board throughout the paper.} This is made possible by three key contributions:
\begin{itemize}
    \item A novel, cache-aware spike transmission algorithm allows linear scaling with network size in the face of millions of neurons that do not fit into cache.
    \item Our parallelization scheme distributes both computations and storage across multiple GPUs.
    \item A simple neuron partitioning strategy achieves perfect load balancing in practice albeit being completely static.
\end{itemize}
The result is a SNN simulator that makes it possible to run ten different experiments of 100K iterations on a \mbox{24B-synapse} model spanning 8 GPUs in the same time it takes competing simulators to create a single network.

%% file: 2_related.tex
\section{Related Work}
As noted by Tuckwell~\cite{tuckwell1988introduction}, early attempts to mathematically model the function of biological neurons can be traced all the way back to the beginning of the 20th century~\cite{lapique1907recherches}. Research on efficiently simulating SNNs has a long history~\cite{Pelayo1997,Mattia2000} and is still showing ever-increasing interest~\cite{VanDerVlag2019,Mozafari2019,bautembach2020,Panagiotou2020,Qu2020,Tian2021}.
It covers areas such as improving the biological fidelity~\cite{Chou2018} and numerical stability of methods~\cite{Tian2021} and employing hardware acceleration~\cite{Smaragdos2017, Sripad2018} on such diverse platforms as VLSI~\cite{Pelayo1997}, FPGA~\cite{Sripad2018}, and even super-computers~\cite{Kunkel2014}. Additionally, there have been theoretical advances such as the quantification of the difference between two spike trains~\cite{VanRossum2001}, the exact solution of differential equations that model membrane dynamics~\cite{Rudolph2006}, and an approach on implementing back propagation on SNNs~\cite{Tomlinson1990}.

\subsection{Simulator classifications}
Several useful classifications of SNN simulators can be defined based on their key traits.
One classification regards whether the simulator aims to model the exact behavior of biological neurons~\cite{Ros2006,Schemmel2010,Thibeault2011,Antonietti2016,Chou2018,Lee2018,VanDerVlag2019a}, or is just based on the general principle of spiking neurons. % all the rest, not citing again

Another classification regards the customizability of the simulators. Simulators that allow the user to specify neuron (and synapse) models with custom behavior can be referred to as general~\cite{Goodman2010,Pecevski2014,Knight2018,Stimberg2018,Hazan2018,Yavuz2016,bautembach2020}, in contrast to ones that only simulate fixed types of neurons~\cite{Schemmel2010,Ahmad2018,Chou2018,VanDerVlag2019a}.

The distinction between event- and time-driven simulation arises in the software architecture of the simulator: Time-driven simulators~\cite{Ros2006a,Rudolph2006,Hanuschkin2010,Ahmad2018,Knight2018,bautembach2020} quantize time into fixed deltas and advance the entire network state in lock-step.
In contrast, event-driven simulators~\cite{Pecevski2014,Ros2006a,Rudolph2006,Mattia2000,Delorme2003,Reutimann2002} intentionally leave neurons in varying, past states until it is required by the simulation to advance them, usually because of incoming spikes.
Typically, time-driven simulators are simpler and more efficient per element but very small deltas can become prohibitively slow. Event-driven simulators offer arbitrary time resolution and can potentially make up for their lower \mbox{per-element} performance by being able to skip ahead in time.
Falling somewhere in between are so-called hybrid approaches~\cite{Naveros2015,Naveros2018}.

Lastly, simulators can also be classified according to the employed hardware platform. Common targets include regular 
CPUs~\cite{Reutimann2002,Delorme2003,Ros2006a,Plesser2007,Garrido2014,Pecevski2014,Panagiotou2020}, GPUs~\cite{bautembach2020,Hazan2018,Qu2020,Mozafari2019,Smaragdos2017,Stimberg2018,Sripad2018,Yavuz2016,Kasap2018,Naveros2015,Ahmad2018,Naveros2018,Szynkiewicz2016,Knight2018,VanDerVlag2019,Krishnamani2010,VanDerVlag2019a,Thibeault2011} (with some works notably combining CPU and GPU computations~\cite{Krishnamani2010, Naveros2015,Naveros2018,Smaragdos2017}), as well as custom/specialized hardware~\cite{Ros2006,Pelayo1997, Schemmel2010, Furber2013, Kunkel2014, Sripad2018}.
There is a clear trend in recent works to increasingly adopt specialized hardware, especially GPUs.

The simulator presented here is of the general-purpose, time-driven, GPU-accelerated variety.

\subsection{Parallel Simulators}
Works that aim to harvest the computational resources of multiple GPUs have been presented as early as 2010~\cite{Krishnamani2010}. 
The simulator in~\cite{Krishnamani2010} is notable for its spike synchronization algorithm, pioneering the use of multiple GPUs as well as utilizing the available CPU. Beyond that it is of no use as its network topology and models were hard-coded.
Also note-worthy is the work by Kunkel et al.~\cite{Kunkel2014} which inspired our parallelization scheme and load balancing strategy. In~\cite{Kunkel2014} the authors took the famous NEST simulator~\cite{Plesser2007} and scaled it to a supercomputer comprising some 100\,000 cores with decent results. Nowadays though it is outperformed by a single GPU by a factor of 100$\times$ (compare Fig.~\ref{figure:sim} ``Brunel+'' with Kunkel et al.~\cite[Fig. 9]{Kunkel2014}).

\input{figures/adj}

As immediate competitors we choose two very recent works, closely related to ours: BSim~\cite{Qu2020} and NeuronGPU~\cite{golosio2020new}. Both simulators are time-driven, GPU-accelerated, and general (to varying degrees), allowing the implementation of the same models in all three simulators, and therefore a direct comparison. A comparison with event-driven simulators does not make much sense since each design pursues distinct goals entailing different trade-offs.
BSim is the only other current, published state of the art simulator with multi-GPU support, so a comparison with it is mandatory. NeuronGPU stands out in that it is more focused on accuracy over performance, featuring double precision arithmetic and Runge-Kutta integration. It is interesting to see which performance trade-offs this design choice implies in practice.

%% file: figures/adj.tex
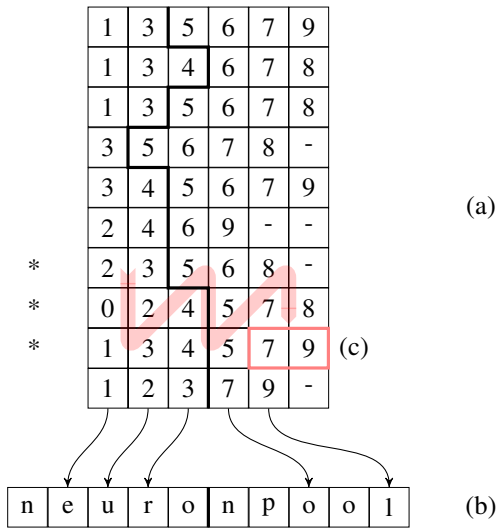
\begin{figure}
    \centering
    \tikzstyle{arr} = [-{stealth'}, thin]
    \tikzstyle{warp} = [line width=2mm, color=red, opacity=0.2, rounded corners]
    \begin{tikzpicture}[node distance = 0.8cm and 0.5cm, x = 1.5em, y = 1.5em]
        % adj list
        \matrix (adj) [
            matrix of nodes,
            nodes = {anchor = center, draw, minimum size = 1.5em, very thin}
        ]{    
            1 & 3 & 5 & 6 & 7 & 9 \\
            1 & 3 & 4 & 6 & 7 & 8 \\
            1 & 3 & 5 & 6 & 7 & 8 \\
            3 & 5 & 6 & 7 & 8 & - \\
            3 & 4 & 5 & 6 & 7 & 9 \\
            2 & 4 & 6 & 9 & - & - \\
            2 & 3 & 5 & 6 & 8 & - \\
            0 & 2 & 4 & 5 & 7 & 8 \\
            1 & 3 & 4 & 5 & 7 & 9 \\
            1 & 2 & 3 & 7 & 9 & - \\
        };
        
        % neuron pool
        \matrix (pool) [
            below = of adj,
            matrix of nodes,
            nodes = {anchor = center, draw, minimum size = 1.5em, very thin}
        ]{
            n & e & u & r & o & n & p & o & o & l\\ 
        };
        
        % connections
        \foreach[count=\src] \dst in {2,3,4,8,10} {
            \draw [arr] (adj-10-\src.south) to [out=270,in=90] (pool-1-\dst.north);
        }
        
        % warp
        \draw [warp, {triangle 90 cap reversed}-] (adj-7-1.east) -- (adj-9-1.north east);
        \draw [warp, -{triangle 90 cap}]
            (adj-9-1.north east) --
            (adj-9-1.east) --
            (adj-7-3.east) --
            (adj-9-3.east) --
            (adj-7-5.east) --
            ($ (adj-9-5.north east) + (0,0.1)$);
        \node (warp) [draw, rectangle, very thick, color=red!50, minimum width=3em, minimum height=1.5em] at (adj-9-5.east){};
            
        % split
        \draw [very thick]
            (adj-1-2.north east) |-
            (adj-2-2.north east) |-
            (adj-2-3.north east) |-
            (adj-3-3.north east) |-
            (adj-3-2.north east) |-
            (adj-4-2.north east) |-
            (adj-4-1.north east) |-
            (adj-5-1.north east) |-
            (adj-5-2.north east) |-
            (adj-8-2.north east) |-
            (adj-8-3.north east) --
            (adj-10-3.south east);
        
        \draw [very thick] (pool-1-5.north east) -- (pool-1-5.south east);
        
        % spikes
        \foreach \i in {7,8,9} {
            \node [left = of adj-\i-1] {*};
        }
        
        % labels
        \node (b) [right = of pool] {(b)};
        \node (a) at (adj -| b) {(a)};
        \node (c) [anchor = west] at (warp.east) {(c)};
    \end{tikzpicture}
    \caption{We store our network's topology in a padded adjacency list (a). Each row stores a neuron's neighbor IDs (b). The synapse pool (not depicted) has the same size as the adjacency list with an implicit 1:1 mapping between the two. CUDA warps (c, here chosen to be 2 threads wide) deliver spikes (*) in a column-wise fashion, keeping recipients close together and improving cache coherency. The thick black line illustrates how the network would be split in two with a pivot of ``4''.}
    \label{figure:adj}
\end{figure}

%% file: 3_method.tex
\section{Method}
For self-containment, we recap the basic data structures and algorithms employed by Spice~\cite{bautembach2020} (Sections~\ref{section:structs&algs}~\&~\ref{section:adjconstr}). We then discuss in depth the proposed spike transmission algorithm and its parallelization (Sections~\ref{section:spikedel}--\ref{section:load}).

\subsection{Data Structures and Algorithms}
\label{section:structs&algs}
Neurons and synapses are stored as a struct of arrays (SoA) which is necessary to enable memory coalescing on GPUs and can dramatically improve achieved memory bandwidth compared to an array of structs (AoS) layout~\cite{nvidia2013}. The user specifies their neuron and synapse formats by inheriting from \lstinline{neuron} and \lstinline{synapse} base classes, for example:
\begin{center}
\lstinline[]!struct myneuron : neuron<float, int> {...}!\\[5pt]
which gets converted into\\[5pt]
\lstinline{tuple<vector<float>, vector<int>>}
\end{center}
via template meta programming. All neurons are stored in a single, global SoA (Fig.~\ref{figure:adj}b) which is as big as the largest neuron model times the total number of neurons. This wastes a few megabytes of memory in exchange for simpler addressing and code. The same holds true for synapses.

The network's topology is stored in a single, padded adjacency list (Fig.~\ref{figure:adj}a). This wastes a few percent of memory in exchange for simpler and faster code  as it requires no offset table and allows us to align rows to \mbox{128-byte} boundaries which leads to a few percent faster memory accesses. Each row's entries are sorted to improve cache coherency. We allocate a synapse pool the size of the adjacency list with an implicit 1:1 mapping between the two (\lstinline{adj[i,j]} corresponds to \lstinline{synapses[i,j]}).

Neuron- and synapse dynamics are specified by implementing callbacks such as \lstinline{onUpdate()} and \lstinline{onReceiveSpike()} which are invoked by the framework on every simulation step. As a result of updating neurons, some of them may spike. Their IDs are inserted into one of $delay$ many spike arrays so they may be delivered on the appropriate simulation step in the future.

\subsection{Adjacency List Construction}
\label{section:adjconstr}
The network's topology is specified through a descriptor of the form $\{\{\textit{range1},\textit{range2},p\},...\}$ which means: ``connect all neurons in $\textit{range1}$ with all neurons in $\textit{range2}$ with probability $p$'' (and so on). The out-degree of neurons in $\textit{range1}$ follows a binomial distribution, which can be approximated by a normal distribution with mean $\mu=|\textit{range2}|*p$ and variance $\sigma^2=|\textit{range2}|*p*(1-p)$. We compute the mean and variance for all neurons, summing them if a neuron is contained in multiple ranges. We then estimate the width of the adjacency list as
\[ \textit{max}\left(\mu_i+3\sqrt{\sigma^2_i}\right), \]
which allows us to pre-allocate it. The layout description is then uploaded to the GPU where it is expanded and from it, the adjacency list is populated.

\input{figures/spikesync}

\subsection{Spike Transmission}
\label{section:spikedel}
Spikes are transmitted by selecting the appropriate spike array $S$ based on the current simulation step and delay. For each spike in $S$, the corresponding row in the adjacency list is selected (Fig.~\ref{figure:adj}*) and the spike is delivered to all neighbors in said row by invoking the aforementioned \lstinline{onReceiveSpike()} callback. This is done by launching $|S|*\textit{width}(\textit{adj})\div32$ CUDA warps where warp $i$ is responsible for delivering spike $i\mod|S|$ to neighbors $\lfloor i\div|S| \rfloor*32$ through $\lfloor i\div|S| \rfloor*32+31$ of neuron $S_{i\mod|S|}$, i.e. column-wise (the neuron is advanced first, then its neighbors, Fig.~\ref{figure:adj}c). Recall that the entries of each row are sorted. By traversing the adjacency list in a column-wise fashion the recipient neurons are statistically expected to be close together, improving our cache hit rate when writing to the global neuron pool. This allows us to scale linearly even when dealing with millions of neurons that do not fit into cache (see Section~\ref{section:results_sparse}). We chose a warp because it is the smallest computational unit that still achieves full memory bandwidth.

\input{figures/spikesync2}

\subsection{Parallelization}
\label{section:parallel}
Our design allows us to split the network across multiple GPUs with \mbox{per-neuron} granularity. Load balancing, the act of deciding which neurons should be assigned to which GPU, will be discussed later. For now let us assume we are given a partition and see how the neuron, synapse, and adjacency data are physically distributed.
We remind ourselves that a network, in its entirety, is defined by the neuron pool, synapse pool, adjacency list, and user callbacks. In order to split such a network into, say, two slices given a pivot one would:
\begin{enumerate}
    \item Split the neuron pool into ranges $[\textit{start}, \textit{pivot})$ and $[\textit{pivot}, \textit{end})$.
    \item Split each row of the adjacency list along the pivot (via binary search).
    \item Split the synapse pool in sync with the adjacency list.
\end{enumerate}
This is illustrated by the thick black line in Fig.~\ref{figure:adj}. Callbacks need not be modified. They will simply be invoked on the new subsets of the original neuron and synapse pools. This split can be performed implicitly during network construction simply by modifying the topology descriptor:
\begin{center}
$\{\textit{range1}, \textit{range2}, p\}$\\[5pt]
would turn into\\[5pt]
$\{\textit{range1}, \textit{range2} \cap [\textit{start}, \textit{pivot}), p\}$ on GPU\textsubscript{0} and\\[5pt]
$\{\textit{range1}, \textit{range2} \cap [\textit{pivot}, \textit{end}), p\}$ on GPU\textsubscript{1}~~~~~~~
\end{center}
Since the network never needs to be instantiated as a whole, the maximum network size grows linearly in the number of GPUs.

It is noted that each half of the original network forms a network in its very own right. It can be simulated on a dedicated GPU independently from the other half and without knowledge of any other GPUs in the system. Each GPU is responsible for delivering \textit{all} spikes to \textit{its} neurons via \textit{its} synapses (similarly to~\cite{Kunkel2014}). Therefore, the only data that need to be exchanged between GPUs are spikes.

\subsection{Spike Synchronization}
Each GPU simulates its network slice in isolation and without knowledge of any other GPUs. A supervisor synchronizes spikes between GPUs using the hierarchical algorithm described in~\cite[pp. 9--18]{nvidia2011} (see Fig.~\ref{figure:spikesync}). The synchronization is performed with a series of \lstinline{cudaMemcpy()} calls so they may overlap with kernel executions, the importance of which will become apparent shortly. Rather than synchronizing spikes after every simulation step, we take advantage of the fact that $delay$ many steps can be executed in a batch: any spikes produced inside one batch will not arrive until the next one (known as ``Timestep Grouping''~\cite{Ahmad2018}). We divide batches into two halves with \( \lfloor delay \div 2 \rfloor \) and \( \lceil delay \div 2 \rceil \) many steps respectively. After the first half completes we asynchronously initiate spike synchronization which then runs concurrently with the second half (Fig.~\ref{figure:spikesync2}). This has two advantages:
\begin{enumerate}
    \item It is faster: By grouping spikes from multiple simulation steps together we cut down on the total number of memory transfers.
    \item We maximize the probability that spike synchronization completes before the next batch begins (which is when the spikes will be needed).
\end{enumerate}
If the simulation step and delay are long enough, we can hence completely hide any overhead introduced by spike synchronization. If, additionally, we can achieve perfect load balancing (all GPUs completing the batch simultaneously), the simulation will scale linearly with the number of GPUs.

\subsection{Load Balancing}
\label{section:load}
The goal of load balancing is to distribute the simulation load across GPUs as evenly as possible so as to avoid the system stalling for a long-running GPU. Ideally, all GPUs would complete every batch at the same time. In our case this means partitioning the neuron pool according to the above criteria. The partitioning of the adjacency list and synapse pool is determined by that of the neuron pool since a GPU must always store the incoming edges to all of its neurons.

PipeDream~\cite{narayanan2019}, which in large part solved multi-GPU backpropagation for conventional ANNs, opts for profiling: Before training, the cost of each layer is measured and layers are then optimally distributed across GPUs. This strategy has several disadvantages and does not map well to SNNs:
\begin{itemize}
    \item The profiling time counts towards the total simulation time. A few seconds of profiling may be negligible in the face of hours of training, but they do add significant overhead in the case of  short-lived SNN simulations.
    \item The thusly obtained neuron costs would only be valid for the profiling period itself. SNNs exhibit very complex dynamic behavior with greatly varying firing patterns \textit{over time}. A partition which results in even load balancing at the beginning of a simulation, may become completely detrimental a few seconds into it.
\end{itemize}

\input{figures/loadbalance}

We opt for a much simpler, completely static solution inspired by~\cite{Kunkel2014}: we split the network into hundreds of \mbox{equal-width} slices and assign them to GPUs in a round-robin fashion (Fig.~\ref{figure:loadbalance}). This averages out any variance or skew present in the neurons' simulation costs. The effectiveness of this strategy depends on the assumption that the number of neurons far outweighs the dynamic range\footnote{the ratio of the largest and smallest values that a certain quantity can assume.} of their costs, which is a necessary condition for any load balancing strategy (with neuron-granularity): If the dynamic range tended towards infinity, the possibility of finding a balanced partition would go to zero. Adversarial cases where every \mbox{$\#GPU$-th} slice happens to be more expensive than the rest are rare and, if they occur, can be alleviated by changing the slice count.

\begin{figure*}
    \centering
    \includegraphics[width=0.68\columnwidth]{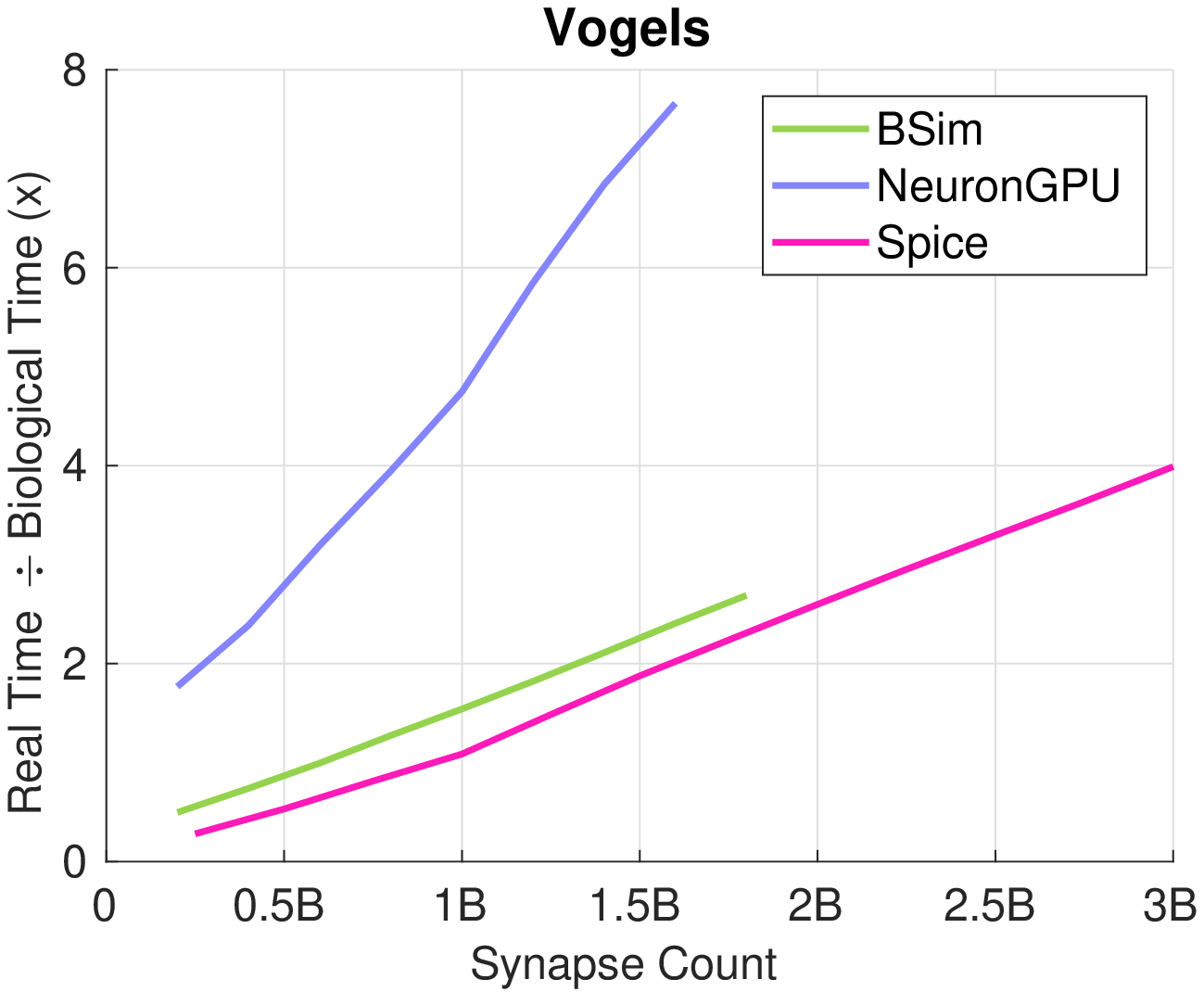}
    \includegraphics[width=0.67\columnwidth, trim=5mm 0 0 0, clip]{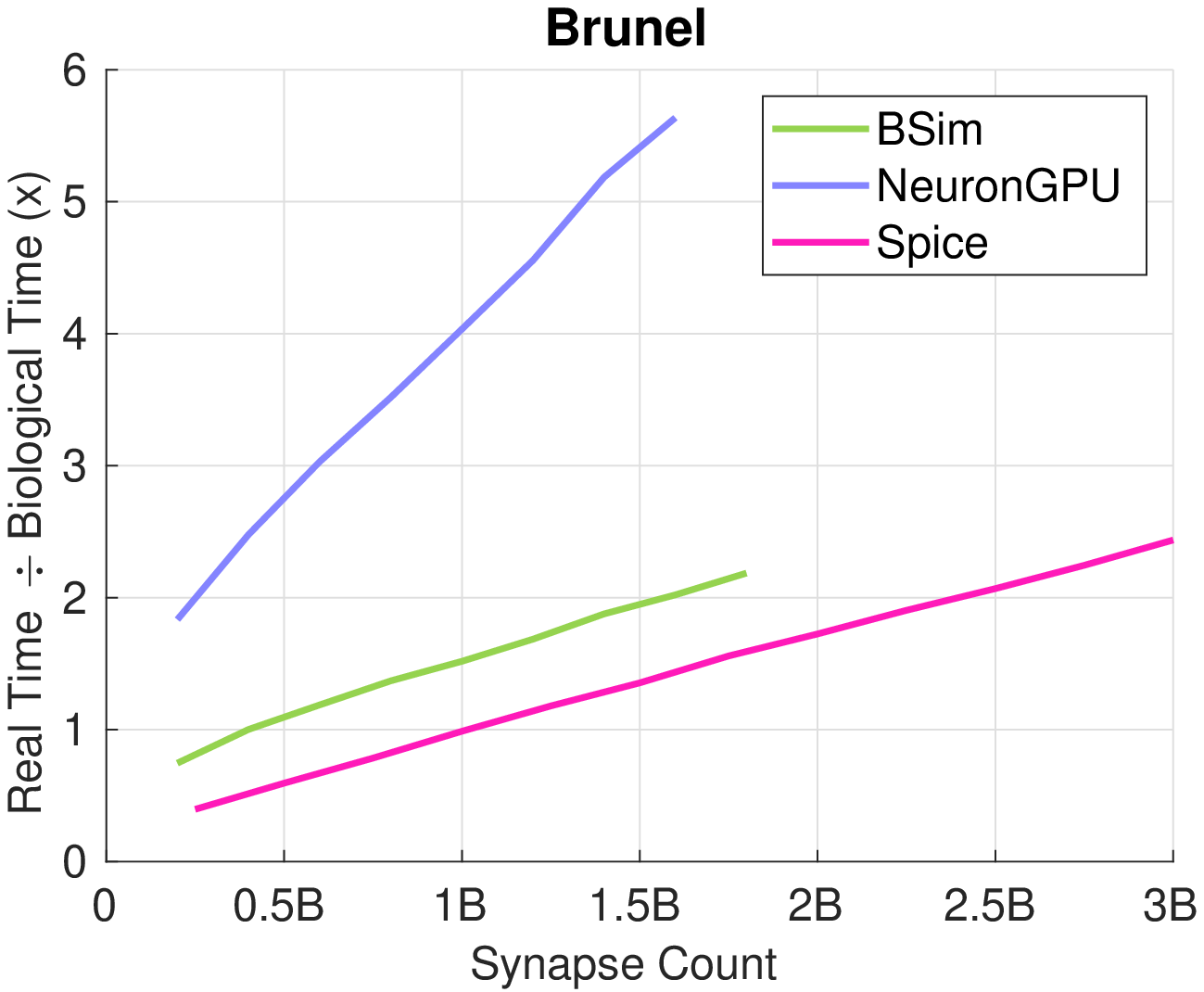}
    \includegraphics[width=0.67\columnwidth, trim=5mm 0 0 0, clip]{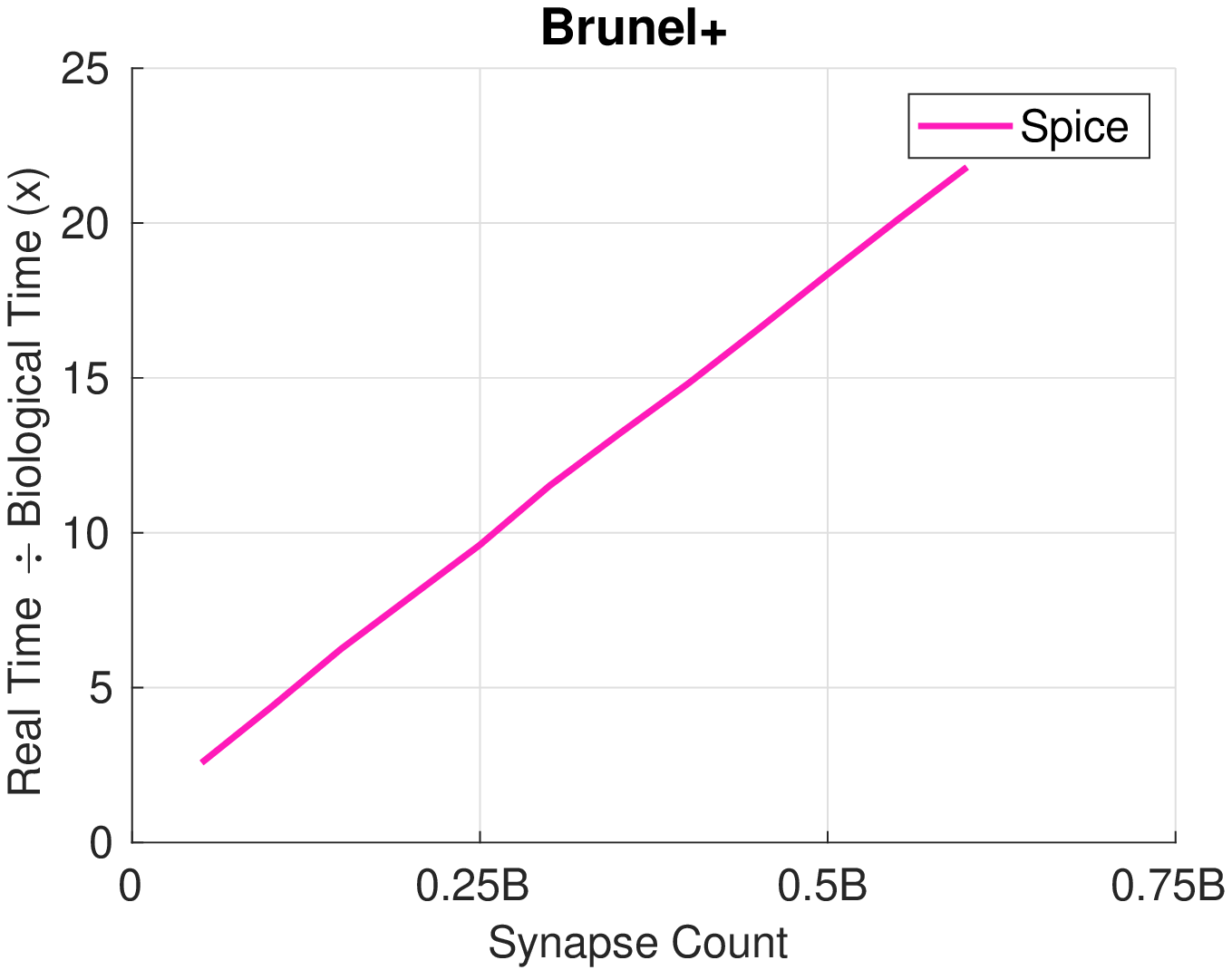}
    \caption{Simulation time as a function of network size. We measure the time it takes to simulate 10 s of biological time for various synapse counts. We report $\textit{wall-clock time}\div\textit{biological time}$.}
    \label{figure:sim}
\end{figure*}
\begin{figure}
    \centering
    \includegraphics[width=0.8\columnwidth]{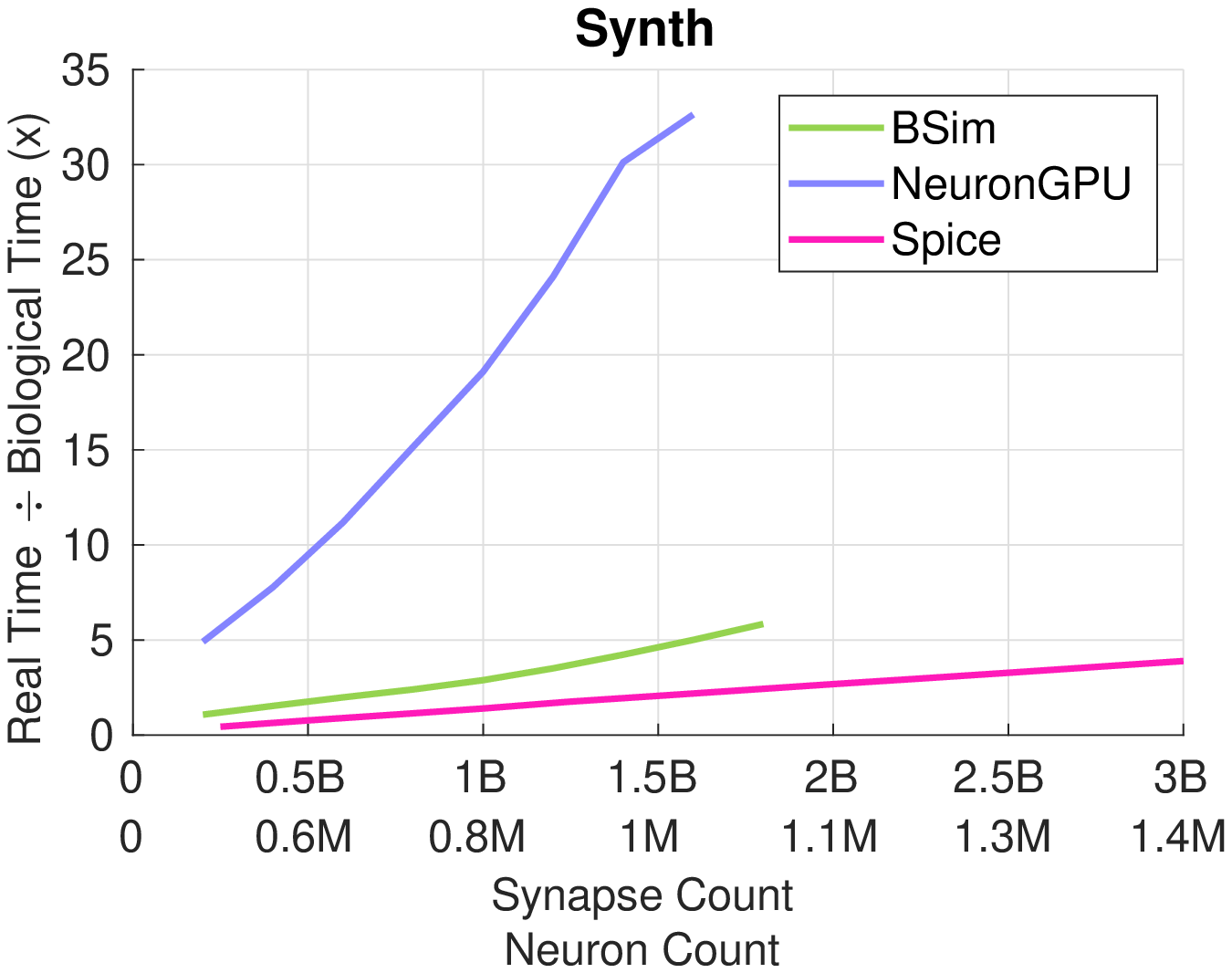}
    \caption{Same as Fig.~\ref{figure:sim} but for the Synth model with $\textit{density}=0.156\%$ and $\textit{activity}=0.5\%$. Neuron counts are provided below synapse counts.}
    \label{figure:synth}
\end{figure}

The split happens analogously to Section~\ref{section:parallel} (albeit with many more pivots) and, once again, can be performed implicitly during network construction. The neuron pool is not physically split but replicated across all GPUs. This wastes a few megabytes of memory but keeps global neuron IDs intact, absolving us from the need for (neuron ID-) translation tables. Each GPU only processes its assigned neuron pool slices by means of simple index manipulation (listing~\ref{listing:loopindex}).

%% file: figures/spikesync.tex
\begin{figure}
    \centering
    \tikzstyle{arr} = [-{stealth'}, thin]
    \begin{tikzpicture}[node distance = 1cm and 1.5cm]
         \matrix (spikes) [
            matrix of nodes,
            column sep = 0.5cm,
            row sep = 0.98cm,
            nodes = {anchor = center, very thin}
        ]{    
            \{a\}       & \{b\}       & \{c\}       & \{d\}\\
            \{a,b\}     & \{b\}       & \{c,d\}     & \{d\}\\
            \{a,b,c,d\} & \{b\}       & \{a,b,c,d\} & \{d\}\\
            \{a,b,c,d\} & \{a,b,c,d\} & \{a,b,c,d\} & \{a,b,c,d\}\\
        };
        
        \draw [arr] (spikes-1-2.south) -- (spikes-2-1.north);
        \draw [arr] (spikes-1-4.south) -- (spikes-2-3.north);
        
        \draw [arr] (spikes-2-1.south) -- (spikes-3-3.north);
        \draw [arr] (spikes-2-3.south) -- (spikes-3-1.north);
        
        \draw [arr] (spikes-3-1.south) -- (spikes-4-2.north);
        \draw [arr] (spikes-3-3.south) -- (spikes-4-4.north);
    \end{tikzpicture}
    \caption{Our spike synchronization algorithm. Depicted are 4 GPUs with their respective spike sets \{a\}, \{b\}, \{c\}, and \{d\}. The goal is for all GPUs to end up with the union of these sets. We hierarchically gather and then distribute the spikes again. Each arrow represents a \lstinline[basicstyle = \footnotesize\ttfamily]{cudaMemcpy()} call. Whenever 2 GPUs end up with half of all the spikes (2nd row, 1st and 3rd GPU), we perform a full duplex sync (crossing arrows) which saves one iteration compared to a strictly hierarchical approach. So long as GPUs are equipped with two copy-engines, all memory transfers between two rows take place concurrently with respect to each other and the ongoing simulation.}
    \label{figure:spikesync}
\end{figure}

%% file: figures/spikesync2.tex
\begin{figure}
    \centering
    
    \tikzstyle{arr} = [-{stealth'}, thin]
    \tikzstyle{abs} = [anchor = north west, minimum height = 1.5em]
    \tikzstyle{rect} = [draw, thin]
    \tikzstyle{spike} = [abs, rect, minimum width = 0.25cm]
    \tikzstyle{sim} = [abs, rect, minimum width = 1cm, rounded corners = 0.1cm]

    \begin{tikzpicture}[node distance = 0, x = 0.25cm, y = -1.95em]
        % grid
        \foreach \x in {5,...,33} {
            \draw [ultra thin, black!40] (\x,0) -- (\x,5);
        }
    
        % GPU labels
        \node [abs] at (0,0) {GPU0:};
        \node [abs] at (0,2.5) {GPU1:};
        
        % sim step timelines
        \def\sat{10};
        \foreach \y in {0,2.5} {
            \node [sim, fill=cyan!\sat] at (5, \y) {};
            \node [sim, fill=cyan!\sat] at (10, \y) {};
            \node [sim, fill=magenta!\sat] at (15, \y) {};
            \node [sim, fill=magenta!\sat] at (20, \y) {};
            \node [sim, fill=cyan!\sat] at (25, \y) {};
            \node [sim, fill=cyan!\sat] at (30, \y) {};
        }
        
        % spike sync timelines
        \node [spike, fill=cyan!\sat] at (15, 1) {$\downarrow$};
        \node [spike, fill=cyan!\sat] at (18, 1) {$\leftrightarrow$};
        \node [spike, fill=cyan!\sat] at (21, 1) {$\uparrow$};
        \node [spike, fill=magenta!\sat] at (25, 1) {$\downarrow$};
        \node [spike, fill=magenta!\sat] at (28, 1) {$\leftrightarrow$};
        \node [spike, fill=magenta!\sat] at (31, 1) {$\uparrow$};
        
        \node [spike, fill=cyan!\sat] at (16, 3.5) {$\downarrow$};
        \node [spike, fill=cyan!\sat] at (18, 3.5) {$\leftrightarrow$};
        \node [spike, fill=cyan!\sat] at (22, 3.5) {$\uparrow$};
        \node [spike, fill=magenta!\sat] at (26, 3.5) {$\downarrow$};
        \node [spike, fill=magenta!\sat] at (28, 3.5) {$\leftrightarrow$};
        \node [spike, fill=magenta!\sat] at (32, 3.5) {$\uparrow$};
        
        % time axis
        \draw [arr] (5,5) -- (33,5) node {\hspace{1em}$t$};
        
        % delay
        \node (delay) [abs, anchor = south] at (14.5, -1) {$delay$};
        \draw (5,0) to [out=45, in=270] (delay.south);
        \draw (24,0) to [out=135, in=270] (delay.south);
        
        % legend
        \node [anchor = north, font = \footnotesize] at (17.5, 5) {
            \begin{tabular}{cl}
                $\downarrow$ & Download spike counts \\
                $\leftrightarrow$ & Synchronize spikes \\
                $\uparrow$ & Upload spike counts \\
            \end{tabular}
        };
        
    \end{tikzpicture}
    \caption{Timeline of a dual-GPU simulation loop with $delay=4$. The top row of each GPU's timeline depicts kernel invocations (simulation steps), the bottom row depicts memory transfers (spike synchronizations). We group the simulation into batches of $delay$ many simulation steps. As soon as half the batch has completed on all GPUs, we download spike counts (required for address calculations), synchronize the spikes (according to Fig.~\ref{figure:spikesync}), and upload the new, total spike counts. Simulation steps and spike transfers of the same color depend on each other's completion.}
    \label{figure:spikesync2}
\end{figure}

%% file: figures/loadbalance.tex
\begin{figure}
    \centering
    \tikzstyle{cell} = [draw, very thin, anchor=north west, minimum width=0.35cm, minimum height=0.35cm]
    \begin{tikzpicture}[node distance = 0, x = 0.35cm, y = -0.35cm]
        % raw data
        \def\adj{{
            {0, 0, 1, 1, 0, 0, 1, 1},
            {0, 0, 1, 1, 0, 0, 1, 1},
            {0, 0, 1, 1, 0, 0, 1, 1},
            {0, 0, 1, 1, 0, 0, 1, 1},
            {0, 0, 1, 0, 0, 0, 1, 2},
            {0, 0, 1, 1, 0, 0, 1, 1},
            {0, 1, 1, 1, 0, 1, 2, 2},
            {0, 0, 1, 1, 0, 0, 1, 2},
            {0, 0, 1, 1, 0, 0, 1, 1},
            {0, 0, 1, 1, 0, 0, 2, 2},
            {0, 0, 1, 1, 0, 1, 2, 2},
            {0, 0, 1, 1, 0, 0, 1, 2}
        }}
        \def\adja{3,3,3,3,4,3,1,3,3,3,2,3}
        \def\adjb{3,3,3,3,1,3,3,2,3,1,2,2}
       
        % adj lists
        \colorlet{color0}{cyan!15}
        \colorlet{color1}{magenta!15}
        \colorlet{color2}{white}
        
        \foreach \y [count=\j from 0] in {0,...,11}
            \foreach \x [count=\i from 0] in {0,...,7} {
                \def\bit{0}
                \pgfmathsetmacro{\bit}{\adj[\j][\i]};
                \node[fill=color\bit, cell] at (\x,\y) {};
            }
        
        \foreach \w [count=\y from 0] in \adja {
            \foreach \x in {0,...,4}
                \node[fill=color2, cell] at ($(\x,\y) + (10,0)$) {};
            \foreach \x in {0,...,\w}
                \node[fill=color0, cell] at ($(\x,\y) + (10,0)$) {};

        }
        
        \foreach \w [count=\y from 0] in \adjb {
            \foreach \x in {0,...,3}
                \node[fill=color2, cell] at ($(\x,\y) + (17,0)$) {};
            \foreach \x in {0,...,\w}
                \node[fill=color1, cell] at ($(\x,\y) + (17,0)$) {};
        }
        
        % neuron pool
        \foreach \x in {0,1,2,6,7,8}
            \node[fill=color0, cell] at ($(\x,0) + (5,14)$) {};
        \foreach \x in {3,4,5,9,10,11}
            \node[fill=color1, cell] at ($(\x,0) + (5,14)$) {};
        
        % labels
        \node at (4,-1) {Adj. List};
        \node at (12.5,-1) {GPU 0};
        \node at (19,-1) {GPU 1};
        \node at (9,6) {$\rightarrow$};
        \node at (16,6) {$+$};
        \node at (23,6) {(a)};
        \node at (23,14.5) {(b)};
        
    \end{tikzpicture}
    \caption{Our load balancing strategy. (a) We split the network into many slices (here 4) and assign them to alternating GPUs, which averages out the variance and skew present in the neurons' simulation costs. (not shown) The synapse pool is split in sync with the adjacency list. (b) The neuron pool is replicated across GPUs as a whole which keeps global neuron IDs intact. Each GPU only processes its assigned neurons though (according to listing~\ref{listing:loopindex}).}
    \label{figure:loadbalance}
\end{figure}

%% file: 4_results.tex
\section{Results}
We compare the performance of our simulator to that of BSim~\cite{Qu2020} and NeuronGPU~\cite{golosio2020new} using three well-established models: Vogels-Abbott~\cite{vogels2005}, Brunel, and Brunel with plasticity~\cite{brunel2000}, detailed in~\cite{Ahmad2018}. We apply a scaling factor to synaptic weights allowing us to vary the network size while maintaining the models' characteristic firing patterns (detailed in~\cite{bautembach2020}). We also use a synthetic model allowing us to vary network topology, size, density, activity (/firing frequency), and delay independently from one another. Unless otherwise noted, all synthetic benchmarks use a single, intra-connected neuron population and a delay of 1. The models will henceforth be referred to as Vogels, Brunel, Brunel+, and Synth.

BSim suffered from race conditions on Volta and newer GPUs which we fixed without negatively affecting performance. NeuronGPU and Spice worked out of the box. We implemented our benchmarks using the C++ API for BSim and Spice, and the Python API for NeuronGPU. All the code used for the experiments can be found at:
\begin{itemize}
    \item \textbf{BSim} \href{https://github.com/denniskb/bsim}{github.com/denniskb/bsim}, forked from master as of Feb 19, 2020.
    \item \textbf{NeuronGPU} \href{https://github.com/denniskb/neurongpu}{github.com/denniskb/neurongpu}, forked from master as of Oct 20, 2020.
    \item \textbf{Spice} \href{https://github.com/denniskb/spice}{github.com/denniskb/spice}, as of Jan 28, 2021.
\end{itemize}

All benchmarks were performed on a Google Cloud VM with an Intel Xeon E5-2699 v3, 8$\times$ Nvidia Tesla V100 16 GB (with P2P access), and 256 GB RAM, running a headless Ubuntu 20 with CUDA 11 and GCC 9.

\begin{figure}
    \centering
    \includegraphics[width=0.75\columnwidth]{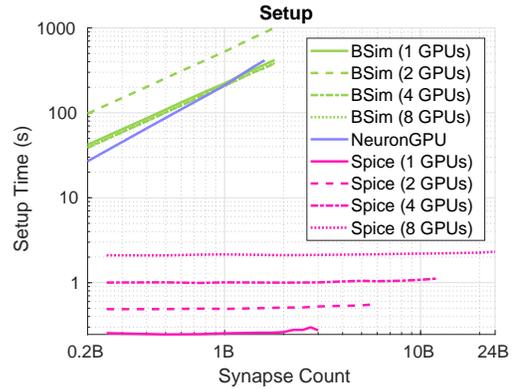}
    \caption{Setup time as a function of network size for the Synth model with $\textit{density}=5\%$. Both axes are logarithmic.}
    \label{figure:setup}
\end{figure}

\subsection{Simulation Time as a Function of Network Size}
\label{section:simtime}
We measure the time it takes to simulate 10 s of biological time for various network sizes (synapse counts). We report $\textit{wall-clock time}\div\textit{biological time}$. All simulators scale close to linearly. While both BSim and NeuronGPU technically support spike-timing-dependent plasticity (STDP), we could not make use of it: BSim did not contain any code samples or documentation illustrating the use of STDP. NeuronGPU does contain a STDP synapse type which unfortunately does not quite reflect the behavior of Brunel+. According to the authors, modifying it ``currently is a task for developers, not for users''. 

\def\figscale{0.71}
\pgfmathsetmacro\gap{1-\figscale}
\begin{figure*}
    \centering
    \includegraphics[width=\figscale\columnwidth]{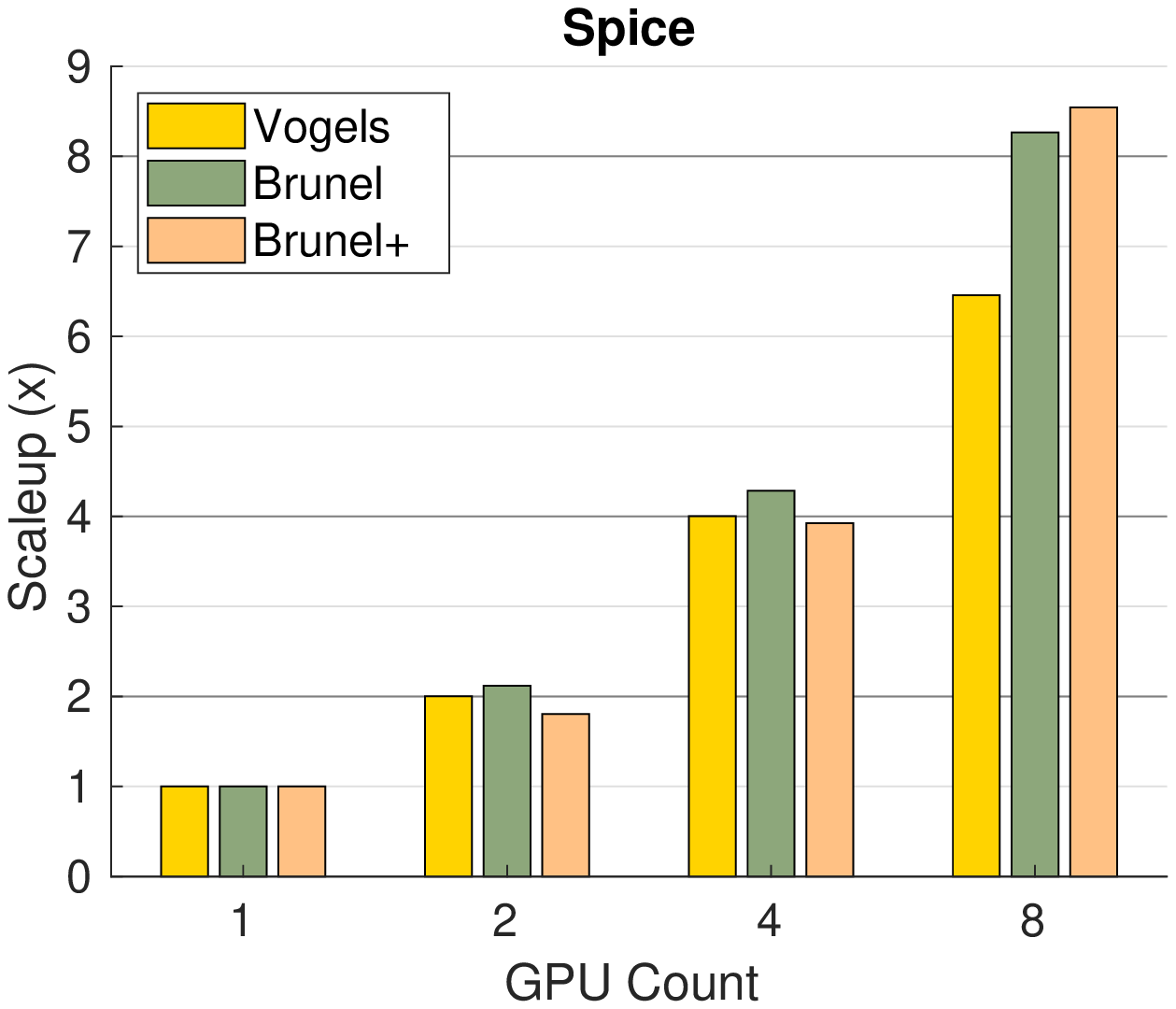}
    \hspace{\gap\columnwidth}
    \includegraphics[width=\figscale\columnwidth]{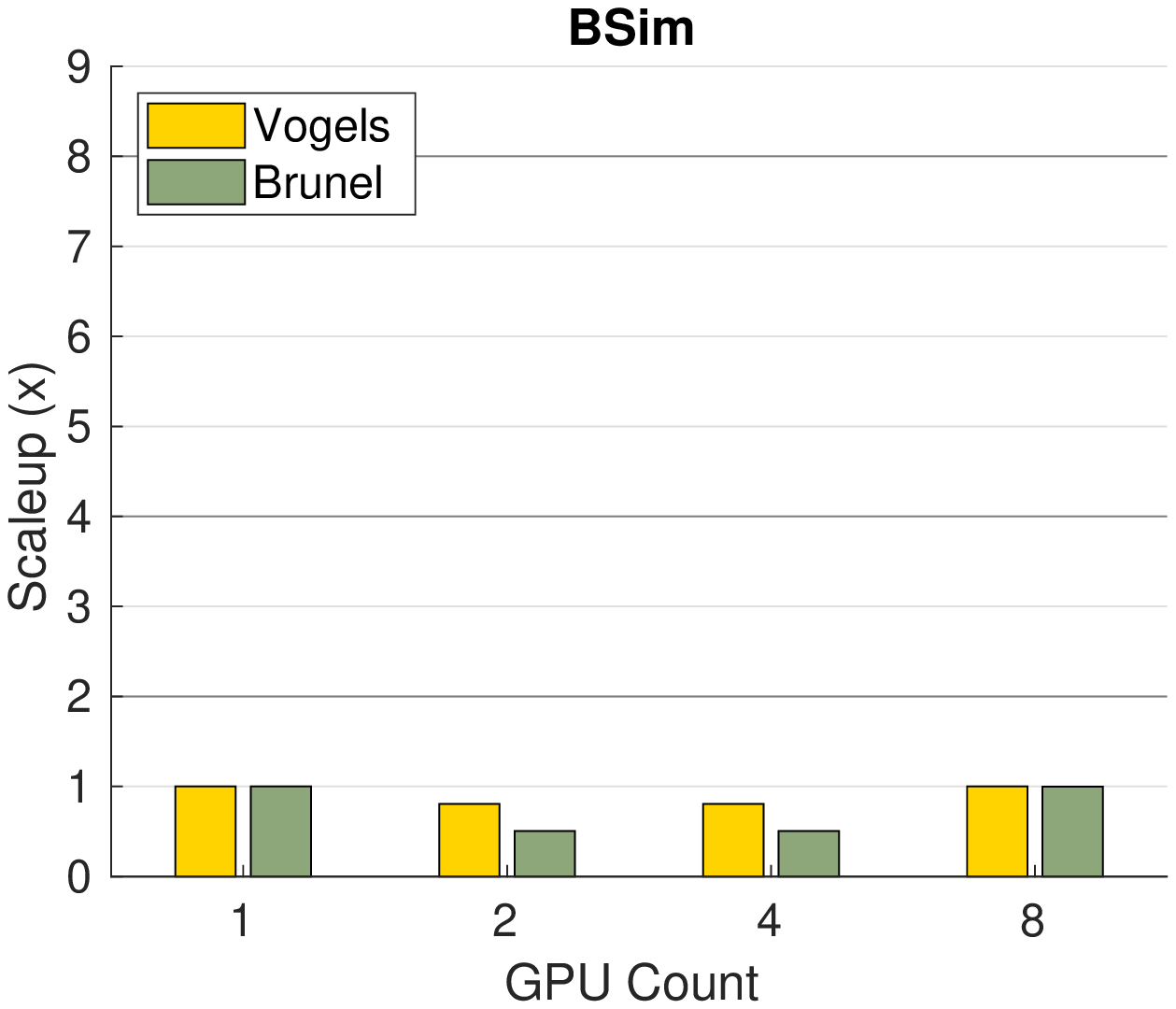}
    \caption{Scaleup as a function of GPU count: How many times larger can we make a model on 2, 4, 8 GPUs while maintaining single-GPU simulation time?}%\vspace{22pt}}
    \label{figure:scaleup}
\end{figure*}
\begin{figure*}
    \centering
    \includegraphics[width=\figscale\columnwidth]{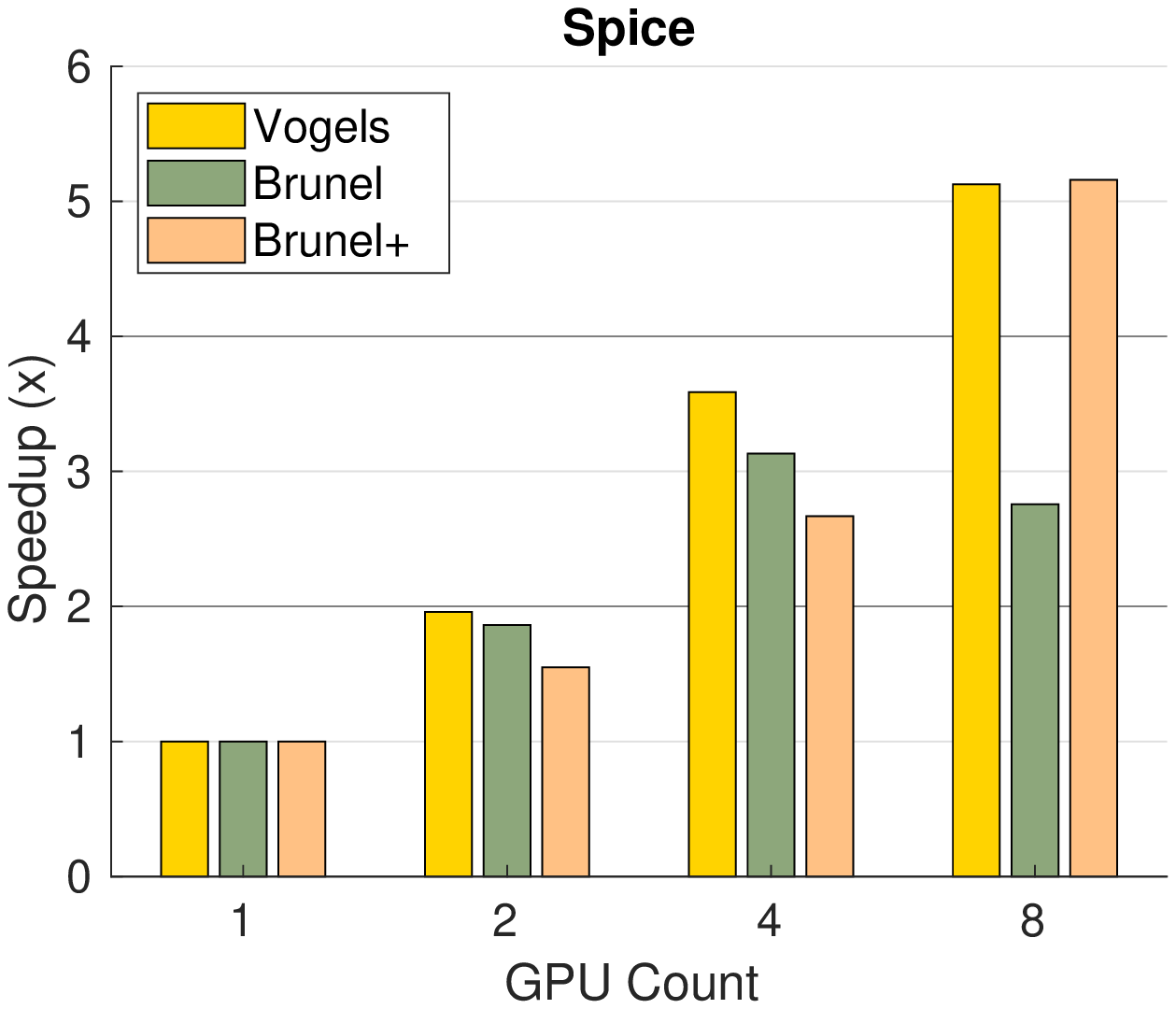}
    \hspace{\gap\columnwidth}
    \includegraphics[width=\figscale\columnwidth]{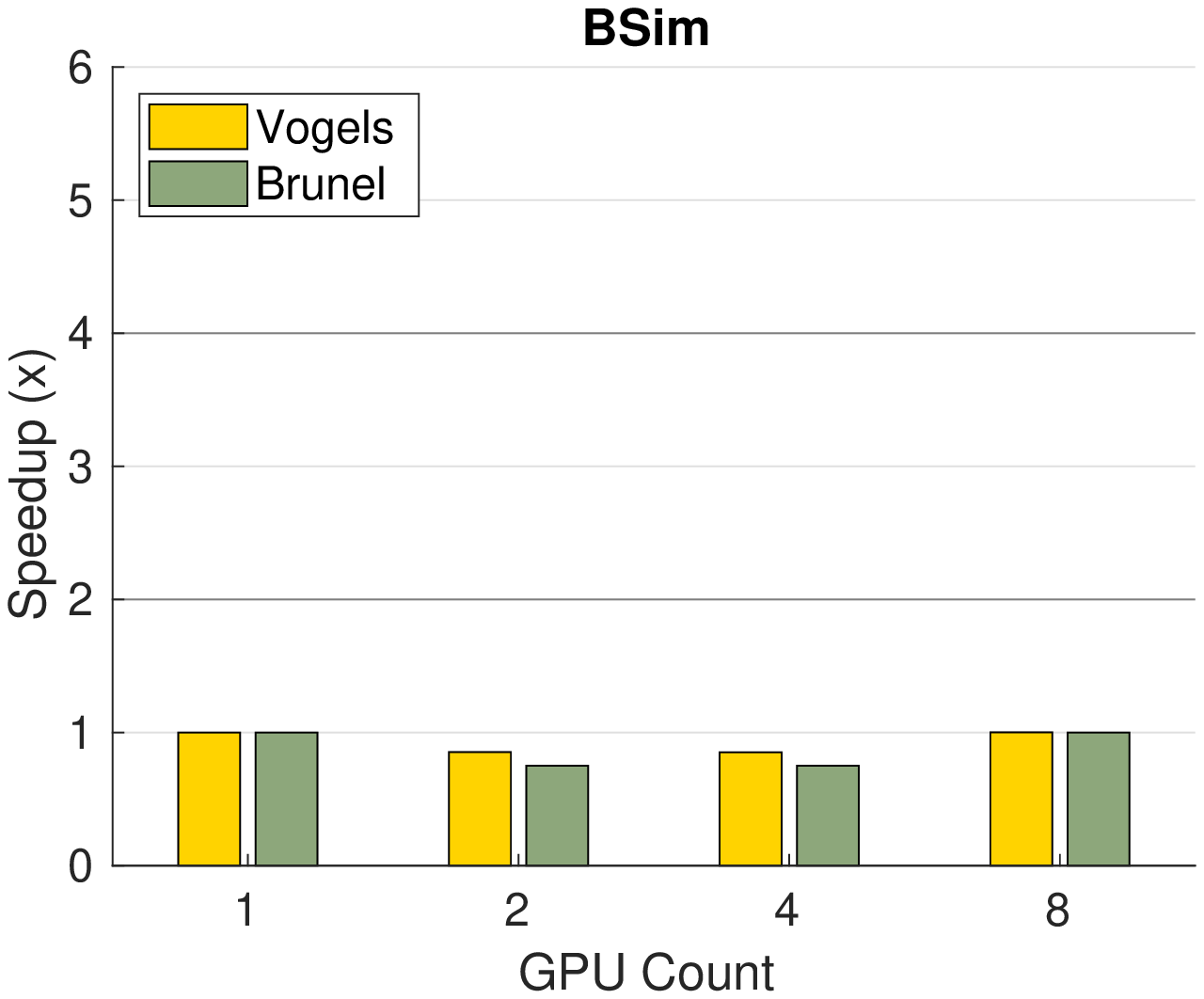}
    \caption{Speedup as a function of GPU count: How much faster does the same model run on 2, 4, 8 GPUs compared to a single GPU?}
    \label{figure:speedup}
\end{figure*}

It is fair noting that NeuronGPU uses double precision arithmetic and ``exact integration''~\cite{Rotter1999} as opposed to single precision arithmetic and Euler integration used by BSim and Spice, which certainly contributes to the performance discrepancy. It is also worth noting that Vogels and Brunel do not perform any faster compared to~\cite{bautembach2020}, in spite of the V100 being \texttildelow25\% faster than the then-used 2080 TI. This is because we are bottlenecked by atomic operations. Brunel+, which is bandwidth-bound on the other hand, does run \texttildelow20\% faster.

\subsection{Simulation Time for Networks with Large Neuron Pools}
\label{section:results_sparse}
Using our Synth model we study the performance of simulators when faced with massive neuron pools that do not fit into cache (Fig.~\ref{figure:synth}). Currently the only way to achieve this is by creating very sparse networks due to limited GPU memory. Both BSim and NeuronGPU seem to exhibit a slight quadratic growth now with the ``bend'' happening at around 800K neurons which coincides with the V100's cache size. Thanks to the employed cache-aware spike transmission algorithm (Section~\ref{section:spikedel}) our simulator continues to scale perfectly linearly. 

\subsection{Setup Time as a Function of Network Size}
Fast setup is important as it allows the user to spend a greater portion of their time on running simulations. It also speeds up parameter exploration and -tuning. We see that Spice's setup is lightning fast because it is performed on the GPU (Fig.~\ref{figure:setup}). We further see that it is virtually constant with respect to network size, meaning it is entirely dominated by spooling up threads and allocating memory---our actual setup kernel generates networks at \texttildelow200M synapses/\textbf{ms}.

\subsection{Multi-GPU Scaling}
Our simulator scales with multiple GPUs in both space and time, allowing one to increase network size while maintaining simulation time (``scaleup'', Fig.~\ref{figure:scaleup}) or to cut down on simulation time while maintaining network size (``speedup'', Fig.~\ref{figure:speedup}). Both scenarios suffer from a natural limit: spike synchronization time grows linearly with the number of GPUs---add enough and any simulation will eventually be bottlenecked by it, leading to sub-linear or even negative scaling. This effect sets in much earlier in the ``speedup'' scenario where \mbox{per-GPU} simulation time goes \textit{down} with the number of GPUs, shortening the spike synchronization window. That is why we consider ``scaleup'' to be the predominant use case of our contribution.

As can be seen, we achieve close to linear scaleup except for Vogels running on 8 GPUs. Vogels has a highly erratic firing pattern alternating between periods during which almost every neuron fires and periods during which few or no neurons fire. While we avoid synchronization altogether if no spikes occur, we still need to download spike counts to determine that no spikes have occurred. In the \mbox{8-GPU} case this takes enough time to delay subsequent simulation steps and prevent linear scaling.

In some cases our simulator even scales \textit{super}-linearly. This is because making a network $n$ times larger and then splitting it into $n$ slices again does not yield the original network. If the original network needed to deliver $S$ spikes to $d$ recipients each, then a slice would need to deliver $S*\sqrt{n}$ spikes to $d\div\sqrt{n}$ recipients each. While the total amount of work stays the same, the trade-offs change.

BSim performs slower on multiple GPUs compared to a single GPU in our benchmarks. According to the authors, BSim ``works well for huge networks with a large number of populations but does not perform well with small number of populations, especially when the numbers of neurons in each population are imbalanced''. Our benchmarks fall into this category. On models that do favor BSim, the authors report a speedup of 1.5$\times$--2.3$\times$ using 4 GPUs~\cite[Fig. 7]{Qu2020}.

\subsection{Memory Consumption}
Each simulator's memory usage  is summarized in Table~\ref{table:memory}. BSim and NeuronGPU use twice the memory to store a synapse's adjacency information, compared to Spice. For non-plastic models this limits them to half the network size. For plastic models this difference is negligible as then VRAM consumption is dominated by the size of the synapse pool.

More importantly though, BSim and NeuronGPU use very memory-inefficient, intermediate data structures before compacting and uploading them to the GPU. This results in very high RAM usage (peaking at 200 GB in the case of BSim) which may be prohibitive for some users.

\input{figures/memory}

%% file: figures/memory.tex
\begin{table}
\renewcommand{\arraystretch}{1.3}
\begin{center}
\caption{Comparison of memory consumption by simulator}
\label{table:memory}
\begin{tabular}{|p{2cm}|l|c|c|c|}
    \hline
    \multicolumn{2}{|l|}{\textbf{Simulator}} & \textbf{BSim} & \textbf{NeuronGPU} & \textbf{Spice}\\
    \hline\hline
    \multicolumn{2}{|l|}{Synapse footprint (adj. data)} & 8 bytes & 9 bytes & 4 bytes\\
    \hline
    \multirow{3}{=}{Max. synapse count (V100)} & Vogels & 1.8B & 1.6B & 3.5B\\
    \cline{2-5}
    & Brunel & 1.8B & 1.6B & 3.5B\\
    \cline{2-5}
    & Brunel+ & - & - & 0.8B\\
    \hline
    \multicolumn{2}{|l|}{Peak RAM usage} & 200 GB & 40 GB & 100 MB\\
    \hline
\end{tabular}
\end{center}
\end{table}

%% file: 5_conclusion.tex
\section{Summary and future work}
We presented a SNN simulator that utilizes all available hardware at close to 100\% efficiency, enabling the simulation of larger models in less time than ever before. The employed multi-GPU parallelization-, spike synchronization-, and latency hiding techniques are not tightly coupled to the rest of our pipeline and should be adaptable to a variety of simulators. Merely the strided load balancing strategy might prove difficult to adapt, seeing as most simulator designs lend themselves to a striped approach.

Several areas for future work can be identified. 
First, we would like to add support for \mbox{per-synapse} delays. Since the number of distinct delays is finite (and rather small for most models), an obvious solution would be to store \mbox{per-delay} adjacency lists. Instead of transmitting all spikes from $delay$ steps ago, one would (in a loop) transmit \mbox{1-step} old spikes via the first adjacency list, \mbox{2-steps} old spikes via the second adjacency list, and so on. The total amount of work remains the same and with the help of CUDA Dynamic Parallelism one could also avoid overhead from additional kernel launches. More importantly, this feature would work completely orthogonally to the existing pipeline: each \mbox{per-delay} adjacency list would be split across multiple GPUs just as it is now. We could still hide spike synchronization latency but its efficiency would be determined by the minimum delay in the system.

\input{figures/loopindex}

Typical spike arrays are so small that the spike synchronization time is entirely dominated by CUDA API overhead (\lstinline{cudaMemcpy()} calls), effectively serializing operations that ought to run concurrently. It is worth investigating whether the spike synchronization could be carried out more efficiently by the CPU instead, especially for eight or more GPUs.

Another interesting research direction is procedural connectivity~\cite{Knight2021}. The authors only store the \textit{parameters} used to create the network and then generate adjacency data on the fly. Perhaps procedural generation could be used for other aspects of a SNN model, too.

%% file: figures/loopindex.tex
\begin{lstlisting}[label = listing:loopindex, caption={The basic structure of our neuron update kernel. The neuron pool (`neurons') is replicated across all GPUs. Each GPU only processes its assigned slices though, via simple manipulation of the loop index.}, captionpos=b, float, belowskip=0pt]
// TID  = thread ID
// NT   = thread count
// S    = slice width
// NGPU = GPU count
// GID  = GPU ID
// N    = neuron count
void updateNeurons(...) {
  for (int i = TID; ; i += NT) {
    int j = (i / S * NGPU + GID) * S + i % S;
    if (j >= N) return;
    
    onUpdate(neurons[j]);
  }
}
\end{lstlisting}